\documentclass{article}

 \usepackage[final,nonatbib]{neurips_2025}

\usepackage[utf8]{inputenc} 
\usepackage[T1]{fontenc}    
\usepackage{hyperref}       
\usepackage{url}            
\usepackage{booktabs}       
\usepackage{amsfonts}       
\usepackage{nicefrac}       
\usepackage{microtype}      
\usepackage{xcolor}         
\usepackage[numbers]{natbib}
\usepackage{ober}
\usepackage{caption}
\usepackage{subcaption}

\usepackage{placeins}

\title{Is Sequence Information All You Need for Bayesian Optimization of Antibodies?}
\workshoptitle{AI for Science Workshop (NeurIPS 2025)}

\author{%
  Sebastian W. Ober \\
  BigHat Biosciences\\  
  \texttt{sober@bighatbio.com}
\And
Calvin McCarter\\
BigHat Biosciences\\
\And
Aniruddh Raghu\\
BigHat Biosciences\\
\And
Yucen Lily Li\\
NYU\\
\And
Alan N. Amin\\
NYU\\
\And
Andrew Gordon Wilson\\
NYU\\
\And
Hunter Elliott\\
BigHat Biosciences\\
\texttt{helliott@bighatbio.com}
}

\begin{document}

\maketitle

\begin{abstract}
Bayesian optimization is a natural candidate for the engineering of antibody therapeutic properties, which is often iterative and expensive.
However, finding the optimal choice of surrogate model for optimization over the highly structured antibody space is difficult, and may differ depending on the property being optimized.
Moreover, to the best of our knowledge, no prior works have attempted to incorporate structural information into antibody Bayesian optimization.
In this work, we explore different approaches to incorporating structural information into Bayesian optimization, and compare them to a variety of sequence-only approaches on two different antibody properties, binding affinity and stability.
In addition, we propose the use of a protein language model-based ``soft constraint,'' which helps guide the optimization to promising regions of the space.
We find that certain types of structural information improve data efficiency in early optimization rounds for stability, but have equivalent peak performance.
Moreover, when incorporating the protein language model soft constraint we find that the data efficiency gap is diminished for affinity and eliminated for stability, resulting in sequence-only methods that match the performance of structure-based methods, raising questions about the necessity of structure in Bayesian optimization for antibodies.
\end{abstract}

\section{Introduction}

Therapeutic antibodies are an important class of drugs that are rapidly increasing in popularity for the treatment of a wide range of challenging diseases \citep{carter2018next}.
To develop a successful antibody for therapeutic purposes, it is not only necessary that it binds to its target at the desired strength, but it must also have satisfactory ``developability'' properties \citep{jarasch2015developability}: for instance, the antibody must be thermostable, have low hydrophobicity, and express well.
Structure-based diffusion generative models have proven useful in non-iterative antibody engineering \citep{luo2022antigen} and \emph{de novo} antibody design \citep{bennett2024atomically}, yet in general yield antibodies still requiring further refinement to meet the criteria of a valid therapeutic. 
In order to satisfy these criteria, it is common to require numerous rounds of iterative optimization using wet lab experiments.
The relatively small datasets involved and the high cost of wet lab experiments form an ideal setting for a technique such as Bayesian optimization \citep{garnett2023bayesian}, which attempts to optimize these properties in an uncertainty-guided way.
However, successfully applying Bayesian optimization requires a number of difficult design choices, most notably in the choice of surrogate model and acquisition function.

A number of works have attempted to elucidate some of these choices, either by proposing new methods \citep[e.g.,][]{stanton2022accelerating, gruver2023protein,amin2025bayesian}, or by benchmarking different methods on relevant problems \citep{gessner2024active}.
However, to the best of our knowledge, none of these works have attempted to incorporate \emph{structural} information, which has been shown to be beneficial for non-Bayesian iterative antibody optimization \citep{raghu2025guided}. 
Moreover, it is unclear which antibody properties structural information might be relevant for: developability properties intrinsic to the antibody, or binding properties, a function of the antibody-antigen (Ab-Ag) interaction. 
The latter is especially uncertain in the common scenario considered here, where the binding pose of the antibody to its target is not known, and cannot be reliably predicted  \citep{abramson2024accurate}.

In this work, therefore, we attempt to understand how best to incorporate structural information, in which situations it is helpful, and how it compares to sequence-only approaches in the low-data regime that is most amenable to Bayesian optimization.
In particular, we aim to address the following questions:
\begin{itemize}
    \item How can we incorporate structural information into Bayesian optimization surrogate models?
    \item For what tasks does structural information aid optimization?
    \item Does incorporating antibody-specific structural models boost performance over general protein models?
    \item Finally, is structural information \emph{necessary} for good optimization performance?
\end{itemize}
Finally, we propose a novel (to the best of our knowledge) means of incorporating sequence-only prior information through a protein language model ``soft constraint,'' which we use to help elucidate answers to these questions.

\section{Background}
Bayesian optimization \citep[BO][]{garnett2023bayesian} is a powerful uncertainty-aware framework for the optimization of expensive black-box functions.
Given potentially noisy observations $\Data = \cb{\x_i, y_i}_{i=1}^N$ of the target function $g: \mathcal{X} \rightarrow \Reals$, BO constructs an uncertainty-aware surrogate model $f$ from the data.
The surrogate model is then used in tandem with an acquisition function, $a: \mathcal{X} \rightarrow \Reals$, which is used to determine which point to query next. 
The choice of query point is typically achieved by balancing exploration and exploitation using the surrogate model's uncertainty.
The point and its acquired value are then appended to $\Data$, and the process is repeated until we exhaust our evaluation budget of the expensive function.

The choice of surrogate model and acquisition function is crucial to the success of BO.
The most common model class for BO is the Gaussian process \citep[GP][]{rasmussen2003gaussian}.
Using a GP, we model the data as $y_i = f(\x_i) + \epsilon_i$, with $\epsilon_i \sim \N{0, \sigma^2}$, where we have placed a GP prior on $f$, given by $f \sim \mathcal{GP}\b{\mu, k}$.
Note that a GP is defined entirely by its mean function, $\mu : \mathcal{X} \rightarrow \Reals$, and its kernel function, $k : \mathcal{X} \times \Xspace \rightarrow \Reals$.

Perhaps the most popular acquisition function for BO is expected improvement \citep[EI][]{Mockus1978}, due to its simple closed-form expression for GPs and its strong empirical performance.
However, EI by itself is inadequate for handling constraints on the search space, which is often a requirement for real-world use: for instance, for antibody development we need to optimize its properties subject to the constraint that it expresses well, which has to be learned.
In order to address this, \cite{gardner2014bayesian} derived the expected contrained improvement acquisition function,
\begin{align}\label{eq:constrained-BO}
    a\b{\x} = \mathrm{PF}\b{\x} \mathrm{EI}\b{\x},
\end{align}
where $\mathrm{PF}\b{\cdot}$ is the learned probability of feasibility.

\section{Methods}

We now turn to describing the methods that we use for our comparison of sequence-only and structure-based Bayesian optimization of antibodies.
We provide further mathematical details in Appendix~\ref{app:deets}, and provide a comprehensive list of evaluated methods in App. Table \ref{tab:method-summary}.

\subsection{Pareto-aware batch Bayesian optimization}

In typical wet lab setups, we often wish to acquire data in batches, ranging in size from tens to hundreds of molecules.
To enable these large batches, we use the recent qHSRI approach of \cite{binois2025portfolio}.
This approach finds the Pareto front of predicted mean and standard deviations, and uses a Sharpe ratio ``portfolio approach'' to decide which variants to select for the batch, based off the predicted mean-standard deviation hypervolume.

As opposed to BO over continuous spaces, where powerful multi-start gradient-based optimizers are typically used, discrete spaces pose an additional challenge when it comes to acquisition function optimization.
We therefore follow \cite{moss2020boss} and use a genetic algorithm.
However, as we wish to find the predicted mean-standard deviation Pareto front, we turn to NSGA-II \citep{deb2002fast}, a genetic algorithm designed specifically for Pareto-aware optimization.
We refer the reader to App.~\ref{app:qhsri-deets} for more details on the qHSRI acquisition function.

\subsection{Sequence-only methods}

Inspired by recent work \citep{gessner2024active, griffiths2024gauche}, we wish to include a strong sequence-only GP baseline.
Following these works, we implement a Tanimoto kernel GP \citep{tripp2023tanimoto} with a constant mean function, using one-hot encodings as our baseline method, which we denote \textbf{OneHot-T}.
Additionally, we use an encoding derived from the BLOSUM-62 matrix \citep{henikoff1992amino}, which was shown to be effective in \cite{gessner2024active} (\textbf{BLO-T}).
Finally, we use the mean-pooled embeddings derived from the ESM-2 650M model \citep{lin2022language} as inputs to a Mat\'{e}rn-5/2 kernel GP (\textbf{ESM-M}).

As an additional baseline, we also compare against LaMBO \citep{stanton2022accelerating}, a complementary sequence-only approach that acquires sequences by using the latent space of a learned denoising autoencoder.
Note that LaMBO has its own acquisition function, and is therefore the only method that we consider that does not use qHSRI acquisition.

\subsection{Incorporating structural information}

One of the main goals of our study is understanding in which scenarios structural information is useful in antibody property optimization.
To that end, we investigate a number of options for incorporating such information.
The first, and simplest, option is to use predicted structures (we assume we do not have access to ground truth structures, especially for each newly designed antibody we wish to consider) as direct inputs to a Mat\'{e}rn-5/2 kernel GP.
To facilitate this, we predict the structure using IgFold \citep{ruffolo2023fast}, align the structure to the predicted parental structure, and extract (and flatten) the alpha-carbon coordinates as the encoding.
We denote this model by \textbf{IgFold-M}. Note that this incorporates \textit{only} 3D structural information without the identities of the specific amino acids comprising the protein and without reference to other known proteins or antibodies.
We additionally consider two approaches that combine sequence and structure information. 
First, we concatenate these features to the ESM-2 features we use above as inputs, which we denote by \textbf{IgFold-ESM-M}.
Second, we combine the IgFold-M kernel with the BLO-T kernel as a weighted sum kernel, which we denote by \textbf{IgFold-BLO-T}.

Finally, we consider the recently-proposed Kermut GP model \citep{groth2024kermut}, which attempts to combine structural information with sequence information.
This model uses a parental structure (which we predict using IgFold), and ProteinMPNN \citep{dauparas2022robust} predictions from that structure alongside a sequence-only kernel to build a sequence-structure kernel for GP modeling. Note that this form of structural information implicitly compares to other known proteins, as ProteinMPNN was trained on a large set of known structures to estimate probabilities of specific amino acids given a structural motif.
Additionally, Kermut uses zero-shot protein language model (pLM) predictions (defaulting to ESM-2 predictions) as part of its prior mean function.
Note that we make some modifications to Kermut, which we ablate in App.~\ref{app:kermut-ablation}.
We denote the final improved model by \textbf{Kermut-T}.

\subsection{Incorporating antibody-specific information}

Due to the particularities of antibodies (for instance, their combination of highly conserved frameworks with hyper-variable complementarity determining regions), we consider incorporating antibody-specific information into the Kermut model.
In particular, we consider replacing the ProteinMPNN predictions with those from AbMPNN \citep{dreyer2023inverse}, with the resulting method denoted as \textbf{AbMPNN-Kermut-T}.
We explore various additional modifications in App.~\ref{app:kermut-imp}.

\subsection{Incorporating sequence prior information as a soft constraint}

One major issue with many of the above pure-GP methods is that they do not incorporate prior information about the likelihood of an antibody sequence.
Given the exponential size of antibody space, it would be wasteful to explore unlikely mutations.
Moreover, without any information on the likelihood of certain mutations, BO will likely explore highly ``unnatural'' mutations that would cause the protein to fail to express or fold, leading to an inability to obtain meaningful property data.\footnote{Indeed, we have observed in our in-house experiments that pure-GP BO \emph{does} explore mutations that cause e.g., expression failures.}
To address this, and inspired by work on constrained BO from Eq.~\eqref{eq:constrained-BO}, we propose incorporating pLM predicted probabilities as a ``soft constraint'' on the acquisition function:
\begin{align}\label{eq:soft-constraint}
    a_{\mathrm{pLM}}\b{\x} = \psub[\mathrm{pLM}]{\x} a\b{\x},
\end{align}
where $\psub[\mathrm{pLM}]{\cdot}$ is the pLM's (pseudo)-likelihood of a sequence.
In practice, we use the pseudo-likelihood from the Sapiens pLM \citep{prihoda2022biophi} as a lightweight antibody-specific pLM for this purpose.

\section{Experiments}

We perform our experimental evaluation \emph{in silico}, focusing on optimizing binding strength and stability using in-house oracles that are trained using data from a real-world optimization campaign, which we describe further in App.~\ref{app:oracles}.
More specifically, we focus on optimizing an antibody's predicted dissociation constant ($K_D$) and melting temperature ($T_m$), starting with 50 examples taken from the early stages of the campaign.
We perform nine acquisitions, aiming to acquire 80 molecules each iteration.
However, in order to ensure the robustness of the BO algorithm, and to increase the fidelity of our \emph{in silico} evaluation to the real world, we randomly drop 30 molecules each iteration: this could be representative of expression failures or other measurement failures that our oracles might not capture.
Finally, we run each experiment three times, and plot the mean performance with standard error bars.
Our experiment therefore hopefully captures meaningful differences between affinity and developability optimization, while remaining faithful to the low-data regime we wish to understand better.

\subsection{Sequence-only methods}

\begin{figure}[!htb]
    \centering 
    \begin{subfigure}[t]{0.49\textwidth}
        \centering
        \includegraphics[width=\linewidth]{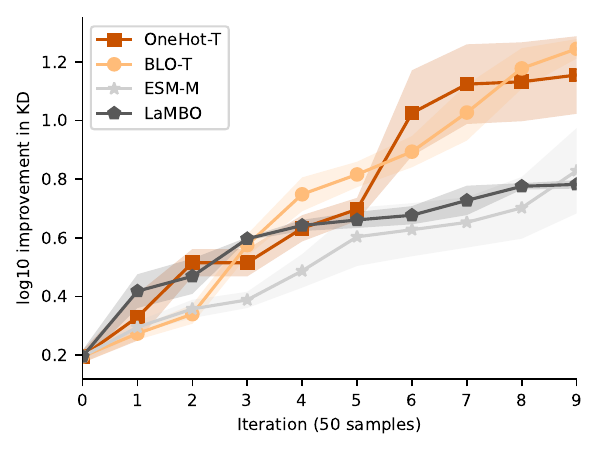}
        \caption{Optimizing for affinity}
        \label{fig:seq_aff}
    \end{subfigure}
    \hfill 
    \begin{subfigure}[t]{0.49\textwidth}
        \centering
        \includegraphics[width=\linewidth]{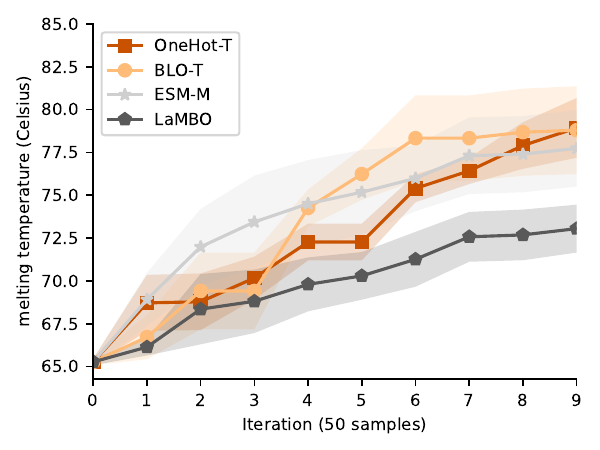}
        \caption{Optimizing for $T_m$}
        \label{fig:seq_tm}
    \end{subfigure}
    \caption{Results on binding affinity $K_D$ and stability $T_m$ for sequence-only approaches. We plot the $\log_{10}$-fold improvement in $K_D$ over the parental sequence for affinity, and the $T_m$ in $^\circ$C for thermostability.}
    \label{fig:sequence}
\end{figure}

We first investigate the performance of sequence-only methods, without the use of any soft constraints.
We compare the OneHot-T, BLO-T, and ESM-M models, along with LaMBO, in Fig.~\ref{fig:sequence}.
We see that for affinity, the Tanimoto kernel models outperform the other methods: note that in the case of ESM-M, this is consistent with the result found in \cite{gessner2024active}.
For $T_m$, we see that while ESM-M has strong initial performance, the Tanimoto kernel models catch up and result in largely equivalent final values.
We note that LaMBO seemingly struggles in both settings, possibly due to the small initial dataset combined with the need to train a denoising autoencoder from scratch.

\subsection{Structure-based methods}

\begin{figure}[!htb]
    \centering 
    \begin{subfigure}[t]{0.49\textwidth}
        \centering
        \includegraphics[width=\linewidth]{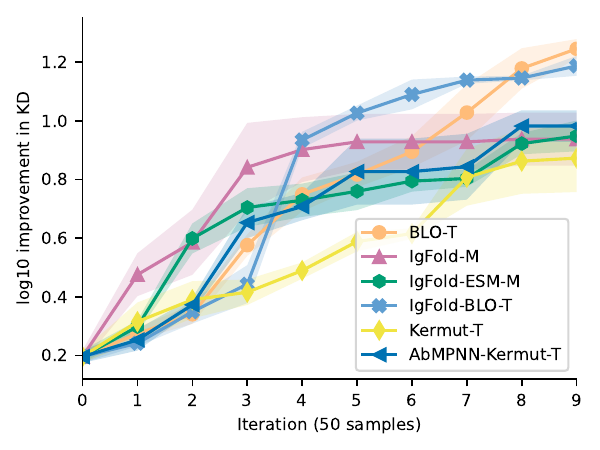} 
        \caption{Optimizing for affinity}
        \label{fig:seq_aff}
    \end{subfigure}
    \hfill 
    \begin{subfigure}[t]{0.49\textwidth}
        \centering
        \includegraphics[width=\linewidth]{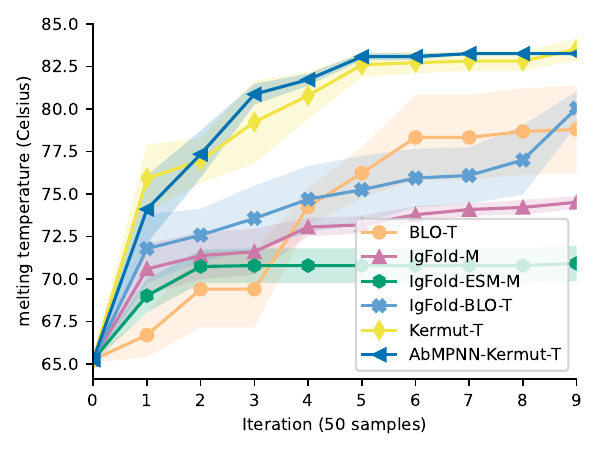}
        \caption{Optimizing for $T_m$}
        \label{fig:seq_tm}
    \end{subfigure}
    \caption{Results on binding affinity $K_D$ and $T_m$ for approaches that incorporate structural information. We also include sequence-only BLO-T for reference.}
    \label{fig:structure}
\end{figure}

We now consider how incorporating structural information affects the optimization of these properties.
We compare the Kermut-T, IgFold-M, IgFold-ESM-M, and IgFold-BLO-T models in Fig.~\ref{fig:structure}.
For reference, we also include the overall best sequence-only model, BLO-T.
We first observe that none of these methods are able to outperform the sequence-only BLO-T approach for affinity, although IgFold-M performs well in the initial iterations.
We hypothesize that this is due to the ability of IgFold-M to more accurately preserve the starting structure, which we explore further in App. \ref{sec:app:rmsd}.
We also see that IgFold-BLO-T, which combines the IgFold-M structure kernel with the BLO-T sequence kernel, results in a middle ground with decent performance in initial iterations and a peak performance similar to BLO-T.

Finally, Kermut-T far outperforms the other methods, including the sequence-only BLO-T model, when it comes to thermostability.
By contrast, its performance on affinity is the worst of the methods considered here.
This points to a fundamental difference in the both features necessary for affinity versus thermostability optimization and the approaches for incorporating structural information.

\FloatBarrier

\subsubsection{Does incorporating antibody-specific information help?}

In Fig.~\ref{fig:structure} we also include comparisons to AbMPNN-Kermut-T, which replaces the ProteinMPNN predictions typically used with Kermut with antibody-specific AbMPNN predictions.
We observe that while there is little change in the performance on thermostability, its performance on affinity is boosted by the change, particularly in its earlier iterations.
These results point to the utility of modality-specific changes.
We consider additional modifications to Kermut in App.~\ref{app:kermut-imp}.

\subsection{Do soft constraints help?}
\begin{figure}
    \centering
    \begin{subfigure}[t]{0.49\textwidth}
        \centering
        \includegraphics[width=\linewidth]{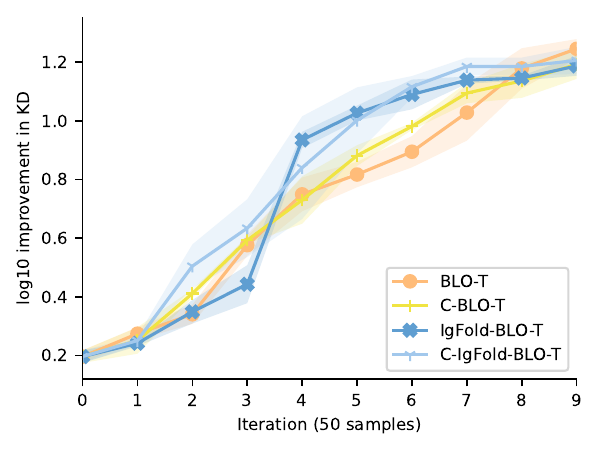}
        \caption{Optimizing for affinity}
        \label{fig:seq_aff}
    \end{subfigure}
    \hfill
    \begin{subfigure}[t]{0.49\textwidth}
        \centering
        \includegraphics[width=\linewidth]{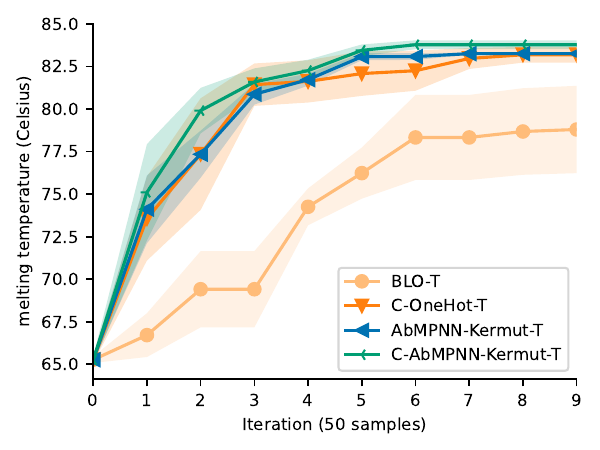}
        \caption{Optimizing for $T_m$}
        \label{fig:seq_tm}
    \end{subfigure}
    \caption{Results on binding affinity $K_D$ and $T_m$ with the inclusion of a pLM-based soft constraint for the best-performing sequence-only and structure-based methods in each setting. }
    \label{fig:pLM}

\end{figure}

We now investigate the effect of incorporating the pLM-based soft constraint of Eq.~\ref{eq:constrained-BO}.
For clarity, we only show the best-performing method for each setting (sequence-only, structure-based, with and without the pLM soft constraint) in Fig.~\ref{fig:pLM}; for full results we refer the reader to App.~\ref{app:plm-full}.
For each method, we use the prefix ``C-'' to denote the constrained version.
We see very little change in affinity optimization from introducing the soft constraint (indeed, the full results show that it can occasionally be detrimental).
However, we see that the sequence-based C-OneHot-T is now able to match the performance of the structure-based methods in optimizing thermostability.
This result dilutes the advantage these structure-based BO methods have to offer over sequence-based methods in this low-data setting, and eliminates it for thermostability.



\section{Discussion \& future work}
In this work, we have investigated the utility of structure-based BO methods versus sequence-only BO methods in the context of both antibody-intrinsic properties such as developability and antibody-target specific properties such as binding affinity. 
We have also assessed the impact of different sequence representations, as well as different forms of structural information: ``purely structural,'' working from 3D coordinates alone (e.g., IgFold-M), and ``statistical,'' estimating likelihoods of structural motifs based on a training set of known proteins (e.g., the ProteinMPNN component of Kermut).

We see that for intrinsic properties such as stability, it is important to have a prior which allows comparison to known proteins, but that this can be either purely sequence based (C-OneHot-T) or of the ``statistical structural'' form (Kermut-T). 
In this case it does not appear to matter whether this information is antibody-specific.

In contrast, for antibody-target pair specific properties, i.e., binding affinity, peak asymptotic performance requires only sequence information, with some benefit from a domain-specific representation (BLO-T is beneficial, though we do not see any benefits from the sequence representations derived from ESM). 
Data efficiency in early iterations is aided by ``purely structural'' information (IgFold-M), which serves primarily to minimize perturbations to the structure of the starting molecule (App. \ref{sec:app:rmsd}). 
Combining the domain-specific sequence representation with this purely structural information provides a compromise in data efficiency and asymptotic performance (IgFold-BLO-T).

However, while some methods do well on both affinity and thermostability, no single method is superior on both, indicating that different features are useful in each case and therefore that an inherent tradeoff exists.
One limitation of our work is that the structural information did not include the target structure in the form of an antibody-antigen complex, as this was not available (as is often the case).
Addressing this in future work might enable structure-based methods to improve their affinity optimization by encouraging mutations that are more likely to perform well for affinity.
Beyond this, the means of incorporating structural information that we considered here were fairly simple: an exciting avenue for future work would be to investigate more sophisticated methods of doing so.
Finally, in future work we plan to validate these methods \emph{in vitro}, and to evaluate whether these observations carry over to other developability properties.

\bibliography{references}
\bibliographystyle{unsrt}

\newpage
\appendix

\section{Related work}
BO for proteins has received an increasing amount of attention in recent years.
\cite{moss2020boss} developed kernels for protein BO and proposed using genetic algorithms for acquisition function optimization.
\cite{gruver2021effective} compare different surrogate models for protein BO.
\cite{stanton2022accelerating,gruver2023protein} propose LaMBO and LaMBO-2, which attempt to perform multi-objective BO by navigating the latent spaces of generative models, with the latter being evaluated on antibody yield and affinity properties.
\cite{benjamins2024bayesian} propose incorporating pLM zero-shot predictions into the prior mean of a GP model for protein BO.
Similar to our work, \cite{gessner2024active} evaluate different surrogate models for BO for antibody affinity; however, their evaluation is limited to sequence-only models, affinity, and single-variant acquisitions, limiting their wider applicability.
More recently, \cite{amin2025bayesian} perform BO using a pLM trained on antibody \emph{families}.
Finally, our pLM soft constraint is similar in formulation to recent works combining generative modeling and Bayesian optimization \citep{moss2025return,yun2025posterior}; however, these works focus on modifying generative modeling sampling to perform BO instead of modifying a standard acquisition function.

\section{Additional methods details}\label{app:deets}
In this section, we give mathematical descriptions of certain methods, where we believe it helpful.
\subsection{Portfolio-based acquisition}\label{app:qhsri-deets}
We briefly describe the qHSRI acquisition function of \cite{binois2025portfolio} that we use for batch acquisition.
Note that we adjust the equations to be suitable for objective function maximization, which better suits our task.
Suppose we have a set of $l$ potential candidates $\x^i$ for the batch, each with predicted mean-standard deviation $\a^i = \b{a_1^i, a_2^i} = \b{m\b{\x_i}, s\b{\x_i}}$.\footnote{Recall that we find the candidates by running a genetic algorithm to find the (approximate) predicted mean-standard deviation Pareto front.}
Then, we define 
\begin{align*}
    r_i &= p_{ii}, \\
    Q_{ij} &= p_{ij} - p_{ii}p_{jj}, \\
    p_{ij} &= \b{\prod_{1 \leq t \leq 2} \b{\min \b{a_t^i, a_t^j} - R_t}} / \b{\prod_{1 \leq t \leq 2} \b{f^*_t - R_t}},
\end{align*}
where $f^*_t$ is the maximum observed value for the dimension in the set of candidates, and $R$ is a lower reference point.
Then, the ``portfolio allocation'' is given by $\z^* \in \sqb{0, 1}^l$ where
\begin{align*}
    \z^* = \argmax_{\z \in \sqb{0, 1}^l}\; h\b{\z} = \frac{\mathbf{r}^\top \z}{\sqrt{\z^\top \mathbf{Q} \z}} \quad \text{s.t.}\quad \sum_{i=1}^l z_i = 1.
\end{align*}
The final batch is given by selecting the candidates with the $q$ highest $z_i$ values.
This acquisition function favors candidates on the predicted mean-standard deviation Pareto front.
Intuitively, this approach can be seen as a method for selecting different values of the $\beta$ parameter in the upper confidence bound acquisition function \citep{srinivas2010gaussian}:
\begin{align*}
    a_\textrm{UCB}\b{\x} = m\b{\x} + \beta \sigma\b{\x},
\end{align*}
where $\m\b{\cdot}$ and $\sigma\b{\cdot}$ are the posterior mean and standard deviation, and forming a batch from optimizing the resulting acquisition functions separately.

In order to modify this with our pLM soft constraint, we multiply the resulting $r_i$ values by the pLM likelihoods.
Note that we modify the genetic algorithm to optimize for pLM probabilities alongside the predicted mean and standard deviations.

\subsection{Kermut}\label{app:kermut-deets}
We briefly describe the salient features of the Kermut model \citep{groth2024kermut} for our analysis.
The Kermut model is a GP model with a particular choice of kernel and mean function.
The kernel is made up of a structure component and a sequence component in a weighted sum:
\begin{align*}
    k\b{\x, \x'} = \pi k_\mathrm{struct}\b{\x, \x'} + (1 - \pi) k_\mathrm{seq}\b{\x, \x'}.
\end{align*}
The sequence kernel is chosen to be a squared exponential kernel on the mean-pooled embeddings from ESM-2.
The structure kernel involves three parts, summed over the effect from each residue that differs from the parental sequence:
\begin{align*}
    k_\mathrm{struct}\b{\x, \x'} = \sum_{i \in M}\sum_{j \in M'} k_\mathrm{struct}^{1}\b{\x_i, \x_j'},
\end{align*}
where $M$ and $M'$ are the sets of mutated residues (with respect to the parental).
$k_\mathrm{struct}^{1}$ itself is made up of three separate kernels:
\begin{align*}
    k_\mathrm{struct}^{1}\b{\x, \x'} = \lambda k_H\b{\x, \x'}k_p\b{\x, \x'}k_d\b{\x, \x'}.
\end{align*}
Here, $\lambda > 0$ is a scalar, $k_H$ represents a Hellinger distance-based kernel on probabilities from an inverse folding model, $k_p$ represents an exponential kernel on the inverse folding probabilities, and $k_d$ is a kernel acting on the physical distance between residues.

The final component of Kermut is the prior mean function, which is chosen to be
\begin{align*}
    m(\x) = \alpha f_0\b{\x} + \beta.
\end{align*}
In this case, $f_0\b{\cdot}$ is chosen to be an ESM-2 zero-shot log-likelihood ratio between variant and wild-type sequences:
\begin{align*}
    f_0\b{\x} = \sum_{i \in M} \log \p{\x_i} - \log \p{\x_i^\textrm{WT}}.
\end{align*}
Note that for multiple mutations, the sum of each individual mutation is taken independently, instead of re-calculating the (pseudo-)log likelihood based on all mutations.

\subsection{Affinity and thermostability oracles}\label{app:oracles}
The oracles we used were derived from ensembles of 10 CARP/ByteNet regressors \citep{yang2024convolutions}.
The affinity ensemble was pretrained on approximately 285,000 sequences from phage display, processed using Next Generation Sequencing (NGS), and fine-tuned on 6,881 sequences with $K_D$ data obtained from Bio-Layer Interferometry (BLI).
Our thermostability ensemble was pretrained on approximately 537,000 sequences from NGS phage display, and 9556 $T_m$ datapoints obtained from NanoDSF. 
The affinity ensemble achieved a test cross-validated Spearman correlation of 0.95, whereas the thermostability ensemble achieved a correlation of 0.72.
Note that for our evaluation, we only use the first model of the ensemble for computational speed.
Finally, we use 159 variants from the early stages of the campaign as the starting set for subsampling for our BO runs.

\section{An ablation study on Kermut}\label{app:kermut-ablation}
Here, we briefly compare different versions of Kermut, motivating the modifications that we take forward.
The first set of modifications, which we denote as Kermut-M, involves a few minor changes, but attempts to keep the overall model largely unchanged.
We first modify the Kermut code to ensure that everything is computed in double precision.
We also disable GPyTorch's $\texttt{fast\_computations}$ settings \citep{gardner2018gpytorch} to ensure exact, Cholesky-based GP inference.
We additionally replace the Adam-based training \citep{kingma2014adam} with BoTorch's $\texttt{fit\_gpytorch\_mll}$ method, which uses L-BFGS \citep{liu1989limited}.
Finally, we notice that the parameterization of the final Kermut kernel is effectively
\begin{align*}
    k\b{\x, \x'} = \sigma_f^2 \pi k_\mathrm{struct}\b{\x, \x'} + (1 - \pi) k_\mathrm{seq}\b{\x, \x'},
\end{align*}
where $\sigma_f^2$ is the GP signal variance, $k_\mathrm{struct}\b{\cdot, \cdot '}$ is the structure kernel, $k_\mathrm{seq}\b{\cdot, \cdot '}$ is the sequence kernel, and $\pi \in \b{0, 1}$ is a weighting.
We hypothesize that a better parameterization is instead
\begin{align*}
    k\b{\x, \x'} = \sigma_f^2 \b{\pi k_\mathrm{struct}\b{\x, \x'} + (1 - \pi) k_\mathrm{seq}\b{\x, \x'}},
\end{align*}
and implement this instead.

Finally, given the relative performance of ESM-based embeddings and one-hot Tanimoto kernels, and given the much greater computational cost of the ESM embeddings, we investigate replacing the default ESM embedding RBF sequence kernel with the one-hot Tanimoto kernel, leading to Kermut-T.
Note that we otherwise retain the modifications used in Kermut-M.

\begin{figure}
    \centering 
    \begin{subfigure}[t]{0.49\textwidth} 
        \centering
        \includegraphics[width=\linewidth]{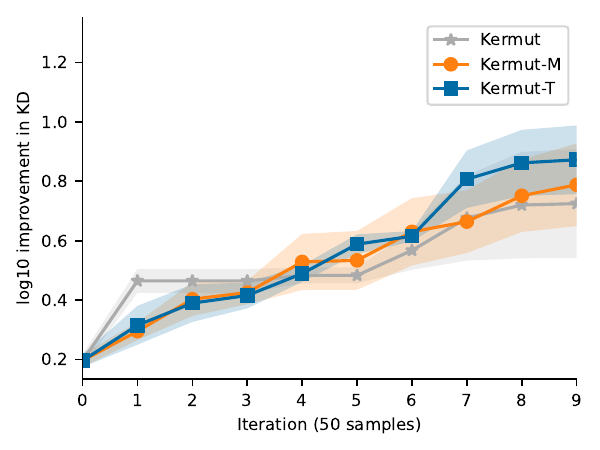} 
        \caption{Optimizing for affinity}
        \label{fig:sub:a}
    \end{subfigure}
    \hfill 
    \begin{subfigure}[t]{0.49\textwidth}
        \centering
        \includegraphics[width=\linewidth]{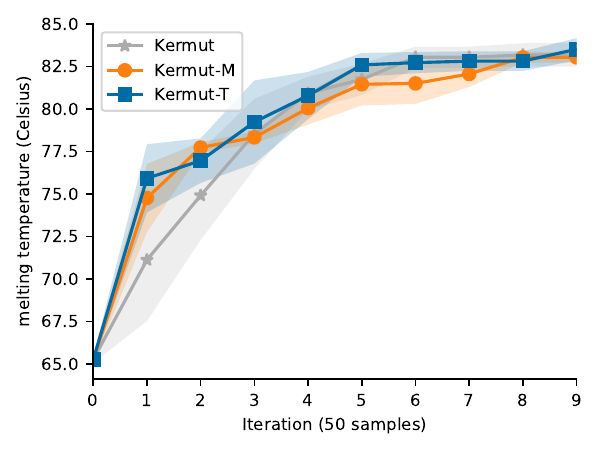}
        \caption{Optimizing for $T_m$}
        \label{fig:sub:b}
    \end{subfigure}
    \caption{Results on binding affinity $K_D$ and $T_m$ for different modifications of Kermut.}
    \label{fig:app:kermut-ablation}

\end{figure}

We plot the results of these experiments on our \emph{in silico} evaluation in Fig.~\ref{fig:app:kermut-ablation}, showing that Kermut-M and Kermut-T match if not outperform the baseline Kermut, justifying our modifications.

\subsection{Further antibody-specific improvements of Kermut}\label{app:kermut-imp}
We now try to ablate different modifications that we could make to Kermut, in hopes of improving its performance.
We focus first on replacing the protein-related deep models involved in Kermut with antibody-specific versions.
To this end, we individually ablate: 
\begin{itemize}
    \item replacing ProteinMPNN with AbMPNN (AbMPNN-Kermut-T);
    \item replacing ESM-2 in the prior mean with a pLM trained on SAbDab (AbSeq-Kermut-T);
    \item replacing the prior mean with a simple learned constant (Const-Kermut-T); and,
    \item given the success of the BLOSUM-based encoding in the sequence-based models, replacing the one hot encoding in the Tanimoto sequence kernel module with the BLOSUM-based encoding (Kermut-BLO-T).
\end{itemize}
Finally, we attempt combining all the antibody-specific models (AbMPNN and SAbDab-trained MLM) and the BLOSUM encoding into a final variation, AbBoth-Kermut-BLO-T.

\begin{figure}
    \centering 
    \begin{subfigure}[t]{0.49\textwidth} 
        \centering
        \includegraphics[width=\linewidth]{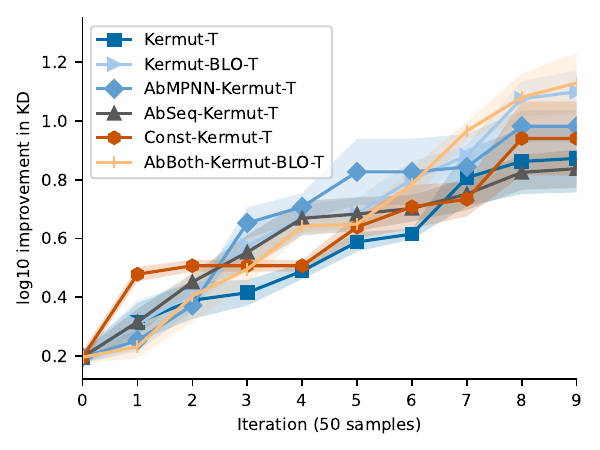} 
        \caption{Optimizing for affinity}
        \label{fig:sub:a}
    \end{subfigure}
    \hfill 
    \begin{subfigure}[t]{0.49\textwidth}
        \centering
        \includegraphics[width=\linewidth]{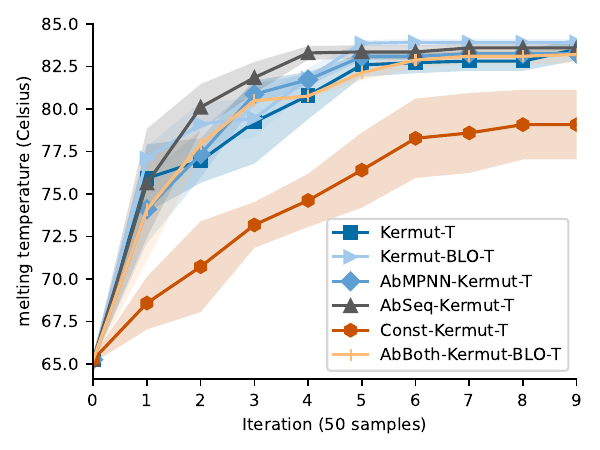}
        \caption{Optimizing for $T_m$}
        \label{fig:sub:b}
    \end{subfigure}
    \caption{Results on binding affinity $K_D$ and $T_m$ for antibody-specific modifications of Kermut.}
    \label{fig:app:kermut-modifications}

\end{figure}

We show these results in Fig.~\ref{fig:app:kermut-modifications}.
These results show that overall, the antibody-specific modifications are helpful.
However, Kermut-B seems to perform the best out of all methods, as the combination of all the antibody-based modifications do not seem to be beneficial together.
Finally, we observe that the use of \emph{some} form of pLM in the prior mean seems beneficial, particularly for $T_m$.

\section{Full soft constraint results}\label{app:plm-full}
\begin{figure}
    \centering
    \begin{subfigure}[t]{0.49\textwidth}
        \centering
        \includegraphics[width=\linewidth]{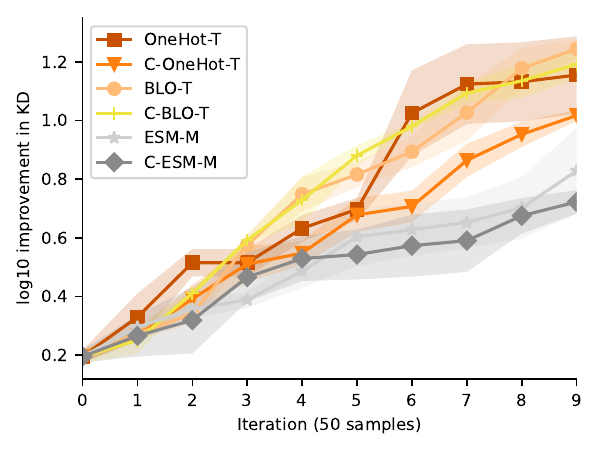}
        \caption{Optimizing for affinity; sequence-only methods}
        \label{fig:seq_aff}
    \end{subfigure}
    \hfill
    \begin{subfigure}[t]{0.49\textwidth}
        \centering
        \includegraphics[width=\linewidth]{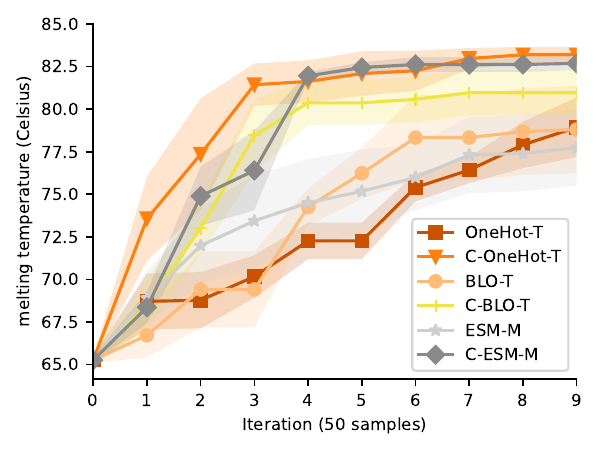}
        \caption{Optimizing for $T_m$; sequence-only methods}
        \label{fig:seq_tm}
    \end{subfigure}

    \begin{subfigure}[t]{0.49\textwidth}
        \centering
        \includegraphics[width=\linewidth]{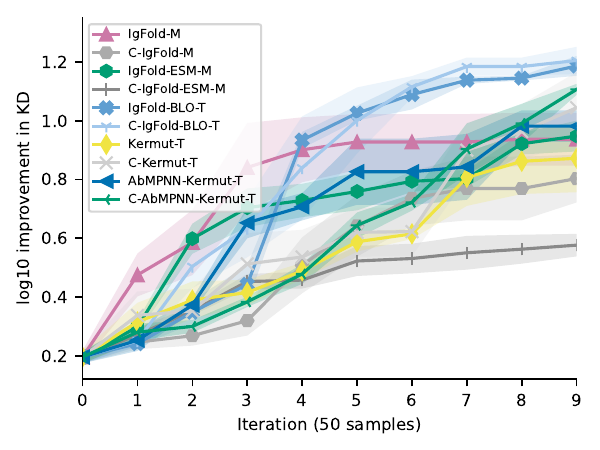} 
        \caption{Optimizing for affinity; structure-based methods}
        \label{fig:struct_aff}
    \end{subfigure}
    \hfill 
    \begin{subfigure}[t]{0.49\textwidth}
        \centering
        \includegraphics[width=\linewidth]{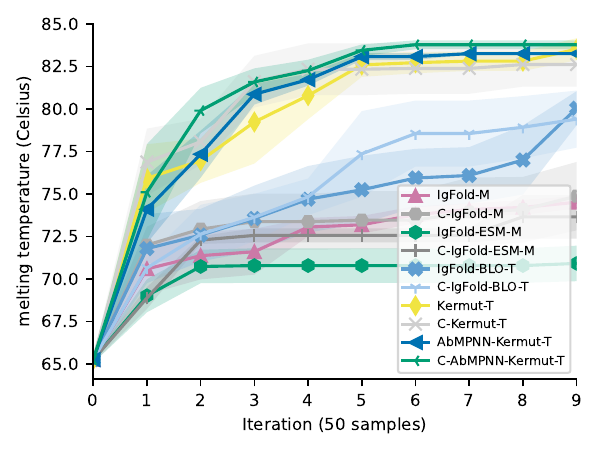}
        \caption{Optimizing for $T_m$; structure-based methods}
        \label{fig:struct_tm}
    \end{subfigure}
    \caption{Results on binding affinity $K_D$ and $T_m$ with the inclusion of a pLM-based soft constraint. Note that we separate out sequence-only (top) and structure-based (bottom) methods for clarity.}
    \label{fig:pLM-full}

\end{figure}

We plot the full results for the soft constraint experiment, separated by sequence-only and structure-based methods, in Fig.~\ref{fig:pLM-full}.

\section{Structural exploration results}
\label{sec:app:rmsd}

\begin{figure}
    \centering 
    \begin{subfigure}[t]{0.49\textwidth} 
        \centering
        \includegraphics[width=\linewidth]{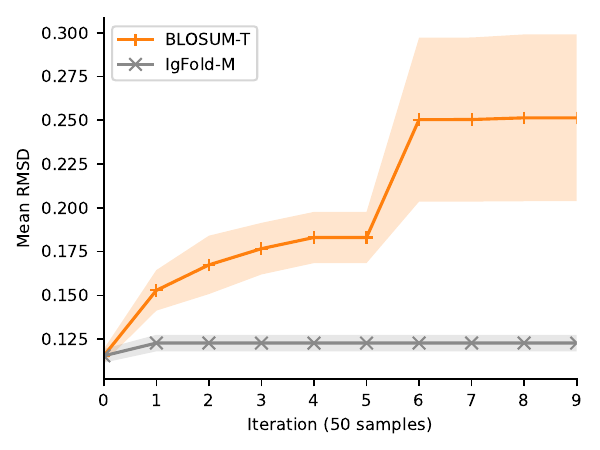} 
        \caption{Optimizing for affinity}
        \label{fig:sub:a}
    \end{subfigure}
    \hfill 
    \begin{subfigure}[t]{0.49\textwidth}
        \centering
        \includegraphics[width=\linewidth]{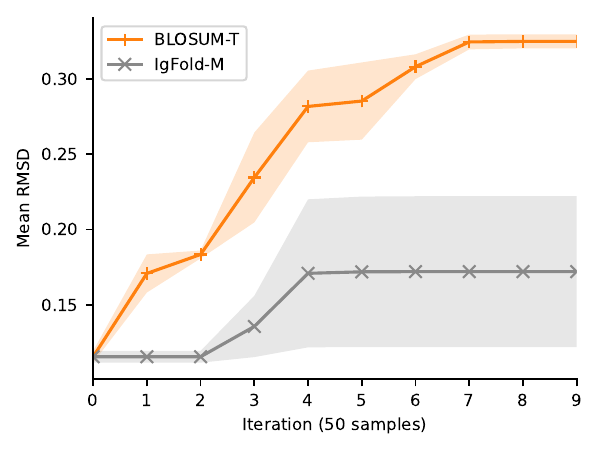}
        \caption{Optimizing for $T_m$}
        \label{fig:sub:b}
    \end{subfigure}
    \caption{Measuring structural exploration from parental when optimizing (a) binding affinity $K_D$ and (b) $T_m$. RMSDs are computed after aligning the IgFold-predicted structures for both the parent and proposed sequences. We compare RMSDs for BLO-T, representing sequence-only optimization, and IgFold-M, representing structure-only optimization.}
    \label{fig:app:rmsd}

\end{figure}

In Figure \ref{fig:app:rmsd} we plot the (predicted) RMSDs between the parental and the proposals over the course of BO iterations.
We see that the sequence-based approach diverges further from the parental in structure-space than the structure-based approach.
This indicates that the structure-based method is better able to hone in on the structural conformation that is most promising for the property at hand.
In particular, for affinity, we do not expect the conformation to change drastically from the parental antibody's conformation when doing iterative optimization.

\section{A summary of our methods}

In Table \ref{tab:method-summary} we summarize the methods evaluated in our work, describing how each of them utilizes sequence and/or structure information.

\begin{table}[]
    \scriptsize
    \centering
    \begin{tabular}{l|c|c|c|c|c|c}
        Method Name & Prior Mean & Seq Rep & Seq Kernel & Struct & Seq-Struct Combo & Constraint \\\hline
        OneHot-T & const & One-hot & Tanimoto & None & NA & None \\
        C-OneHot-T & const & One-hot & Tanimoto & None & NA & Sapiens pl \\
        BLO-T & const & BLOSUM & Tanimoto & None & NA & None \\
        C-BLO-T & const & BLOSUM & Tanimoto & None & NA & Sapiens pl \\
        ESM-M & const & ESM2 emb & Mat\'{e}rn-5/2 & None & NA & None \\
        C-ESM-M & const & ESM2 emb & Mat\'{e}rn-5/2 & None & NA & Sapiens pl \\
        IgFold-M & const & None & NA & IgFold coords & None & None\\
        C-IgFold-M & const & None & NA & IgFold coords & None & Sapiens pl\\
        IgFold-ESM-M & const & ESM2 emb & None & IgFold coords & Concat, Mat\'{e}rn-5/2 & None\\
        C-IgFold-ESM-M & const & ESM2 emb & None & IgFold coords & Concat, Mat\'{e}rn-5/2 & Sapiens pl\\
        IgFold-BLO-T & const & BLOSUM & Tanimoto & IgFold coords & Add kernels & None \\
        C-IgFold-BLO-T & const & BLOSUM & Tanimoto & IgFold coords & Add kernels & Sapiens pl \\
        Kermut-T & ESM2 pll & One-hot & Tanimoto & Composite & Add kernels & None\\
        C-Kermut-T & ESM2 pll & One-hot & Tanimoto & Composite & Add kernels & Sapiens pl\\
        AbMPNN-Kermut-T & ESM2 pll & One-hot & Tanimoto & Composite (Ab) & Add kernels & None\\
        C-AbMPNN-Kermut-T & ESM2 pll & One-hot & Tanimoto & Composite (Ab) & Add kernels & Sapiens pl\\
        Const-Kermut-T & const & One-hot & Tanimoto & Composite & Add kernels & None\\
        AbSeq-Kermut-T & SAbDab pll & One-hot & Tanimoto & Composite & Add kernels & None\\
        Kermut-BLO-T & ESM2 pll & BLOSUM & Tanimoto & Composite & Add kernels & None\\
        AbBoth-Kermut-BLO-T & SAbDab pll & BLOSUM & Tanimoto & Composite (Ab) & Add kernels & None\\     \hline
    \end{tabular}
    \caption{Summary of methods evaluated in this work.}
    \label{tab:method-summary}
\end{table}

\end{document}